\documentclass[10pt,twocolumn,letterpaper]{article}

\usepackage{cvpr}              

\usepackage{graphicx}
\usepackage{amsmath}
\usepackage{amssymb}
\usepackage{booktabs}
\usepackage{multirow}
\usepackage{makecell}

\usepackage[pagebackref,breaklinks,colorlinks]{hyperref}

\usepackage[capitalize]{cleveref}
\crefname{section}{Sec.}{Secs.}
\Crefname{section}{Section}{Sections}
\Crefname{table}{Table}{Tables}
\crefname{table}{Tab.}{Tabs.}

\def\cvprPaperID{***} 
\def\confName{CVPR}
\def\confYear{2023}

\begin{document}

\title{All in Tokens: Unifying Output Space of Visual Tasks via Soft Token}

\author{Jia Ning$^{1,4}$\thanks{Equal Contribution. Jia, Chen, and Zigang are interns at MSRA.}, Chen Li$^{2,4}$\footnotemark[1], Zheng Zhang$^{4}$\footnotemark[1], Zigang Geng$^{3,4}$, Qi Dai$^{4}$, Kun He$^{1}$, Han Hu$^{4}$ \\
\\
{$^1$Huazhong University of Science and Technology} \quad
{$^2$Xi’an Jiaotong University} \\
{$^3$University of Science and Technology of China} \quad
{$^4$Microsoft Research Asia} \\
\small{\texttt{\{t-jianing,t-chenli1,zhez,t-ziganggeng,qid,hanhu\}@microsoft.com}}
\\
\small{\texttt{brooklet60@hust.edu.cn}}
}
\maketitle

\begin{abstract}
Unlike language tasks, where the output space is usually limited to a set of tokens, the output space of visual tasks is more complicated, making it difficult to build a unified visual model for various visual tasks. In this paper, we seek to unify the output space of visual tasks, so that we can also build a unified model for visual tasks. To this end, we demonstrate a single unified model that simultaneously handles two typical visual tasks of instance segmentation and depth estimation, which have discrete/fixed-length and continuous/varied-length outputs, respectively. We propose several new techniques that take into account the particularity of visual tasks: 1) Soft token. We employ soft token to represent the task output. Unlike hard tokens in the common VQ-VAE which are assigned one-hot to discrete codebooks/vocabularies, the soft token is assigned softly to the codebook embeddings. Soft token can improve the accuracy of both the next token inference and decoding of the task output; 2) Mask augmentation. Many visual tasks have corruption, undefined or invalid values in label annotations, i.e., occluded area of depth maps. We show that a mask augmentation technique can greatly benefit these tasks. With these new techniques and other designs, we show that the proposed general-purpose task-solver can perform both instance segmentation and depth estimation well. Particularly, we achieve 0.275 RMSE on the specific task of NYUv2 depth estimation, setting a new record on this benchmark. The general-purpose task-solver, dubbed AiT, is available at \url{https://github.com/SwinTransformer/AiT}. 

\end{abstract}

\section{Introduction}

A unified model for various tasks across modalities is one of the most ambitious goals of artificial intelligence (AI). However, this is challenging due to the diversity and complexity of real-world tasks. Despite the difficulties, large-scale language models in the field of natural language processing (NLP), \eg, GPT-3~\cite{brown2020language}, have demonstrated impressive capabilities as a general-purpose solver for language tasks. Encouraged by the success of NLP, this paper attempts to investigate the possibility of universal models for various computer vision tasks.

Most existing related research for universal computer vision models focuses on the unification of architectures~\cite{jaegle2021perceiver,wang2022image} and pre-training~\cite{baevski2022data2vec}. A few works aim for developing one model for multiple visual tasks, yet they are usually applicable to limited tasks: ~\cite{alayrac2022flamingo} handles only tasks with language as output; CLIP models~\cite{radford2021learning} and the follow-ups~\cite{yu2022coca,yuan2021florence,zhu2022uni} tackle only retrieval and image classification tasks; \cite{chen2022unified} deals with tasks that have describable and sequential outputs. In this paper, we aim for a truly general-purpose solver for various vision tasks. To this end, we first notice that a key obstacle was overlooked and under-explored:  unlike language tasks whose inputs and outputs are represented by language tokens of the same form, the output forms for different computer vision tasks are more diverse. For example, the output of object detection is a set of coordinates and labels; the output of semantic segmentation is a discrete label map; the output of depth estimation is an image with floating-point values; and the output of flow estimation is a vector field.

\begin{figure*}
    \centering
    \includegraphics[width=1\linewidth]{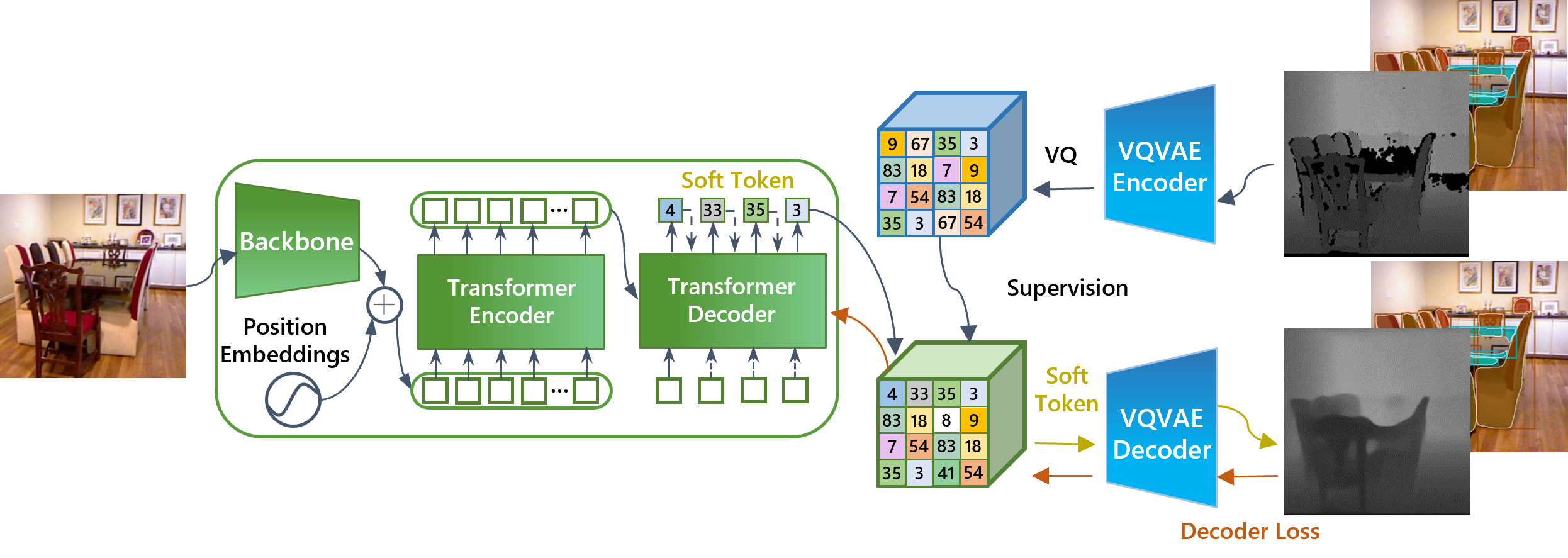}
    \caption{Illustration of our unified framework with two stages. In this framework, various vision task outputs are transferred to discrete token space by a VQ-VAE tokenizer. In this way, discrete or continuous visual tasks can be converted into one discrete classified task,}
    \label{fig:pipeline}
\end{figure*}

In this paper, we address this obstacle by unifying the output space of various visual tasks through a general tokenizer to encode the output space into a set of tokens. Specifically, the proposed framework consists of three components: \emph{tokenizer}, \emph{detokenizer}, and \emph{task solver}, as illustrated in Figure~\ref{fig:pipeline}. The \emph{tokenizer} and \emph{detokenizer} are instantiated by VQ-VAE~\cite{van2017neural}, which encodes the task output into a set of tokens, which are then reconstructed into the original output by the decoder. The \emph{task solver} for different vision tasks is instantiated using an auto-regressive encoder-decoder model. The encoder-decoder model takes images as inputs and predicts a token sequence in a causal mode, which is decoded into the original task-specific output form by the VQ-VAE decoder.

To improve the effectiveness, we propose several new techniques that take into account the particularity of visual tasks. First, we propose soft token instead of hard ones as used in language tasks or by the original visual VQ-VAE framework. The soft token is represented by probability vectors, with each value in a vector denoting the probability belonging to a codebook. When a soft token is fed to the input of the detokenizer or to the input of the next token prediction network, its input embedding is computed by the weighted average of the codebook embeddings based on the probability values. This indicates that the soft token embedding has spanned an interpolable continuous space, which may better represent the visual output, especially when it’s continuous. The continuous nature of the soft token also allows the introduction of an auxiliary loss which learns the task output end-to-end, back from the detokenizer output to the task-solver input. In experiments, we show that the three usages of the soft token can all lead to better results.

Second, we propose a mask augmentation technique to deal with visual tasks which have corrupted, undefined, or invalid values in annotations. The depth estimation is a typical example of such a task, where the occluded area is not defined~\cite{nathan2012nyuv2}, as shown in Figure \ref{fig:aug_recons}. The undefined regions make it difficult to the training of VQ-VAE tokenizers and detokenizers because it does not know what to reconstruct in the undefined area. To solve this problem, we randomly mask several patches of the input depth map when training VQ-VAE. Unlike undefined areas, manually masked patches have ground truth annotations, which help train the VQ-VAE network to be able to recover the ground truth for undefined areas. In experiments, we show this technique to notably improve the accuracy of depth estimation. The mask augmentation technique is also expected to be useful for other visual problems with similar undefined or corrupted annotations, and therefore strengthens the generality of our unified task solver.

Third, we systematically study the impact of architectures in the VQ-VAE model. We find that a very lightweight VQ-VAE with up to 5 convolution layers, 2 residual blocks, and 128 codebooks can serve as a very powerful tokenizer. The number of parameters and computations can be as small as 2M and 0.06G FLOPs. This suggests that the overhead of the VQ-VAE model is small, which improves the practicality of our framework.

We choose two classical visual tasks with varying output spaces to validate our approach: depth
estimation, and instance segmentation, whose output spaces are in formats of the floating-point maps and binary masks, and have fixed size and variable size, respectively. We achieve competitive accuracy on COCO instance segmentation~\cite{lin2014microsoft}, and state-of-the-art accuracy for NYUv2 depth estimation~\cite{nathan2012nyuv2}. The proposed framework and techniques are generic and can be applied to other visual tasks, which will be our future work.

\section{Related Works}

\paragraph{Unified Frameworks in Computer Vision}
Encouraged by the success of T5~\cite{raffel2020exploring}/GPT~\cite{brown2020language} in NLP, the exploration of a single unified model for various tasks in computer vision has emerged. However, most existing works\cite{alayrac2022flamingo, yu2022coca, yuan2021florence,wang2022image,zhu2022uni} are focused on the training algorithm or model architectures. This makes these models either only available as pre-trained models~\cite{yu2022coca,yuan2021florence,wang2022image} or only for VL-related tasks~\cite{alayrac2022flamingo,zhu2022uni}. Perceiver-IO~\cite{jaegle2021perceiver} presents a framework that can process different vision tasks, and it adopts the learned positional encoding or Fourier feature to unify the output of different tasks.

Very recently, Pix2SeqV2~\cite{chen2022unified} proposed to unify different vision tasks into tokens and the tokens in Pix2SeqV2 need to be designed manually for different tasks. For example, they use the polygon to represent the instance segmentation.

The most related works with ours are UViM~\cite{kolesnikov2022uvim} and Unified-IO~\cite{lu2022unified}. They also adopt VQ-GAN/VQ-VAE as a general tokenizer/detokenizer and an auto-regressive Transformer encoder/decoder to solve different tasks. However, they are direct applications of these techniques without an in-depth consideration of the particularity of visual problems.

Our study, while concurrently started, takes more in-depth consideration of the particularity of visual problems. We propose techniques of soft token and mask augmentation, which prove beneficial generally for visual tasks or a part of them. We also extensively investigate the architecture of the VQ-VAE, which shows that this part can be made very light-weight, and thus make the framework more practical.

\paragraph{Vector-Quantization}
Discretized token output space is widely used in generative models, such as DALL-E\cite{ramesh2022hierarchical}, VQGAN\cite{esser2021taming}, and VQ-Diffusion\cite{gu2022vector}, to represent high-dimensional complex data. Models like VQ-VAE\cite{van2017neural}, dVAE\cite{bao2021beit} define a discrete latent space with the encoder-decoder architecture and a fixed size of codebook, The input is mapped to the discrete tokens of the codebook. 
We adopt the VQ-VAE framework to build our token space, and our soft token approach that treats the token space as a continuous one instead of the original discrete one expands the usage of VQ-VAE.

\paragraph{Monocular Depth Estimation}
Monocular depth estimation is a fundamental task for 3D perception. Deep learning dominates the depth estimation since Eigen \etal~\cite{Eigen14} introduces it into the depth task. The follow-up works include proposing powerful network~\cite{Lee19bts, Ranftl21DPT, Li22depthformer}, designing novel augmentation~\cite{Kim22GLP, ishii21cutdepth}, making use of the geometric constraints~\cite{qi18geonet, Yin19}, exploring pairwise relationship~\cite{zoran15rel, chen16sdp, Lee19rel}, combing with conditional random field~\cite{liu15dcnf, xu18str, yuan22newcrf}.

Some works~\cite{Nekrasov19, Zhang19PAP, wang20sdcdepth, Ranftl21DPT, invpt2022} combine depth estimation with other tasks, such as semantic segmentation, and edge estimation. However, they design different heads and loss functions for different tasks respectively. There are also some methods~\cite{Fu18dorn, Diaz19softlabels, Bhat21adabins, Bhat22localbins, li22binsformer} discretizing the continuous depth and cast the depth estimation as a per-pixel classification task. Our approach represents the depth maps as a set of tokens, and unifies it with other visual tasks, \ie instance segmentation, in a unified network structure and output space. 
More importantly, we show that a general framework for various visual tasks can achieve state-of-the-art accuracy on the NYUv2 depth estimation task.

\paragraph{Instance Segmentation} 
Instance segmentation aims to predict the segments of each instance. There are many works~\cite{he2017mask,yang2020dense,xie2020polarmask} studying how to represent the masks. For example, MaskRCNN~\cite{he2017mask} used a binary mask, Dense RepPoints~\cite{yang2020dense} adopts a set of deformable points to represent the segments, and PolarMask~\cite{xie2020polarmask} models the segments by polygons. 
While their representation is specific to instance segmentation, we model the instance segmentation by a set of discrete (soft) tokens, which is more general for visual representations.

\section{Framework}
The goal of this work is to unify the output space of visual tasks into discrete tokens and to build a single model that can handle different tasks simultaneously. In this section, we present the framework to achieve this goal, which is shown in Figure.~\ref{fig:pipeline}. The framework consists of three modules, a \emph{tokenizer} that encodes the task output to the discrete tokens, a \emph{detokenizer} that decodes tokens to the task output, and a \emph{task-solver} that predicts tokens from images. In our approach, the encoder and decoder of VQ-VAE are used as the tokenizer and detokenizer, and the task-solver is implemented by an auto-regressive encoder-decoder model. During training, task annotations are first mapped by the tokenizer as discrete tokens and used as supervision to train the task-solver. In inference, the tokens predicted by the task-solver are decoded by the detokenizer into task outputs.

\subsection{Tokenizer and Detokenizer}
VQ-VAE is an encoder-decoder model with a set of latent codes. It was originally proposed to learn discrete representation for natural images. In this work, we use its encoder and decoder as the tokenizer and detokenizer. In training, the input image is encoded as a set of contiguous embeddings, and these embeddings are assigned to their nearest latent codes. In the decoder, the corresponding codes are used as inputs instead of contiguous embeddings and then decoded into the image. By minimizing the reconstruction loss between the input image and decoded images, the encoder, decoder and latent codes can be trained.

Since we adopt discrete tokens as targets in the task-solver, the accuracy of reconstruction has an upper-bound on the performance of the whole framework. In addition, both training and inference of task-solver require the tokenizer and detokenizer, the fast inference speed is also desired. 
The original network architecture of VQ-VAE is designed for natural images, which have more complex textures and colors than the task output, making it not the optimal design for us. Therefore, we have exhaustively studied the effects of VQ-VAE with different design choices in our framework. 

Typical reconstruction losses (\eg $l$-1 loss, MSE loss, \etc) cannot directly reflect the realistic performance of the task, we use standard evaluation metrics for different tasks to measure VQ-VAE. As shown in Table \ref{tab:codebook_solver} and Table \ref{tab:dim_solver}. We found that a very lightweight VQ-VAE can achieve promising results in depth estimation and instance segmentation. 

Since the input value ranges for depth estimation and instance segmentation are different. We train two VQ-VAE models for two tasks separately, and they have similar architecture. For depth estimation, the encoder consists of 5 convolution layers (kernel size is 3 and stride is 2) and follows 2 residual blocks. The output feature map has a downsample ratio of 32, and the channel dimension is progressively increased from 16 to 256. The architecture of the decoder is symmetrical to the encoder, only replacing the convolution layers with the deconvolution layer. For instance segmentation, we reduce the convolution and deconvolution of the encoder and decoder from 5 layers to 4 layers and keep all others the same. Subsequently, the downsample ratio is changed to 16.

In addition, compared to the standard VQ-VAE usually adopts a large codebook size (\eg 8192), our codebook size can be reduced to 128. We note that though larger codebook size consistently improves VQ-VAE reconstruction ability, they show no difference when applied to task-solver (see Table \ref{tab:codebook_VQ-VAE} and Table \ref{tab:codebook_solver}). There are two speculations on the effectiveness of a small codebook: 1) the output space of depth and segmentation is simple, without the need for a large codebook; 2) the large codebook may increase the learning difficulty for task-solver.

\subsection{Task-solver}
The task-solver is an auto-regressive encoder-decoder network. The encoder is a Swin Transformer with 6 additional standard transformer blocks, each block consists of a self-attention and an FFN. The decoder has 6 blocks, each block consists of a self-attention, a cross-attention, and an FFN. The architecture we used is similar to \cite{chen2021pix2seq}.

Given an input image, the encoder is first applied to learn a generic representation of all tasks. Then, based on the given \emph{task token}, the decoder is used to predict a token sequence in an auto-regressive manner. For different tasks, we customize their sequence formats, as described in the following:

\paragraph{Depth Estimation}
The task token of depth estimation is denoted as \texttt{[DEP]}. It has a straightforward format, which is a token sequence of length $\frac{HW}{32^2}$, where $H$ and $W$ are the height and width of input images, respectively.

\begin{figure}
    \centering
    \includegraphics[width=1\linewidth]{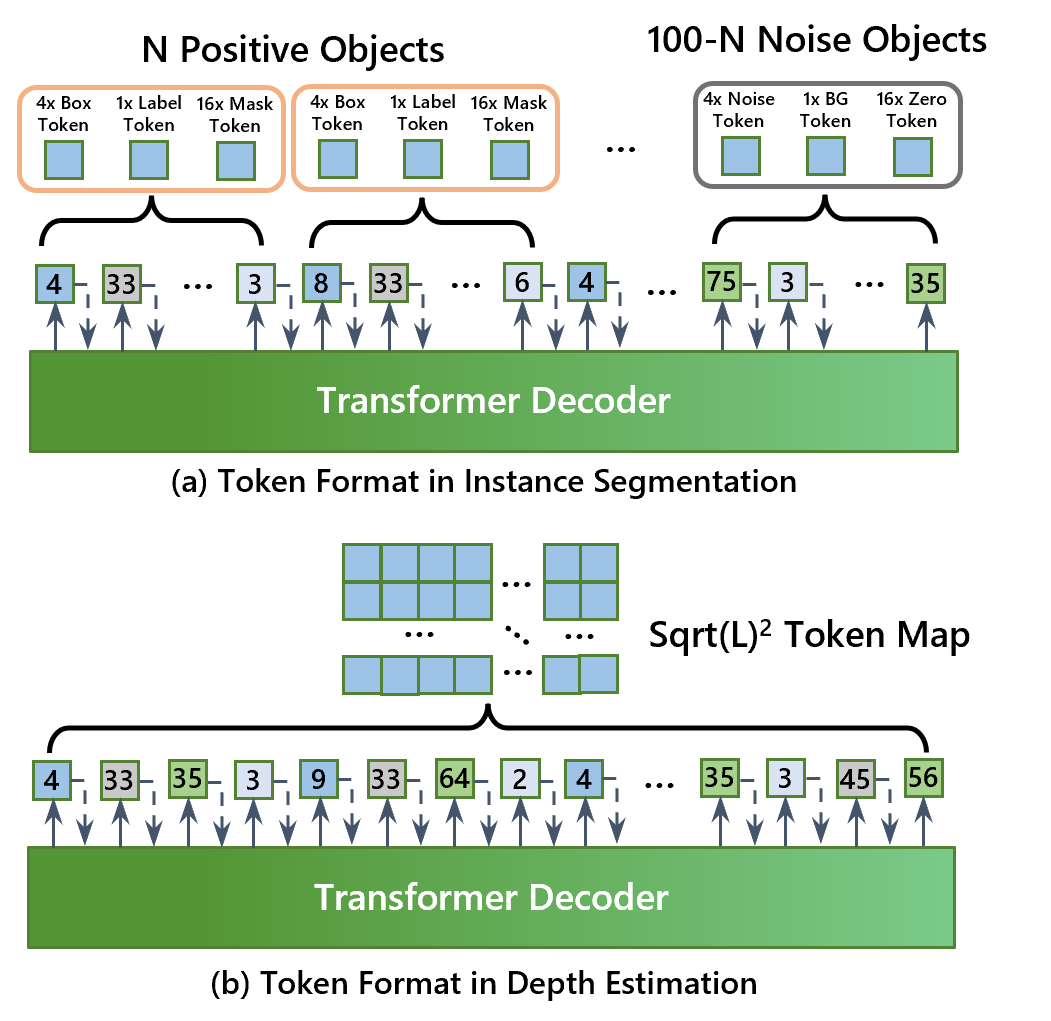}
    \caption{Illustration of instance segmentation and depth estimation token format. (a) We organize every object by 21 tokens. For positive objects, this format includes 4 box, 1 label and 16 mask tokens. While for noise tokens, 4 noise, 1 background and 16 zero tokens are employed; (b) For dense tasks like depth estimation, we treat every patch a token to form a token map.}
    \label{fig:insseg_seq}
\end{figure}

\paragraph{Instance Segmentation} 
The sequence format of instance segmentation is more complicated than depth estimation, which consists of three parts: bounding box coordinates, class of bounding box, and a binary mask. 
We follow the practice of Pix2Seq~\cite{chen2021pix2seq} for representing coordinates and the class of a bounding box. The coordinates are manually quantized into 2000 bins, and different classes are represented by different tokens (including a background class, \eg COCO dataset has 81 class tokens in total). For the binary mask, we use $4\times4$ tokens to represent one mask. It is worth noting that since the computational complexity of auto-regressive is proportional to the square of the output sequence length, we have to use very few tokens to represent the mask. Nevertheless, benefitting from our powerful detokenizer, these tokens can be decoded to a $64\times64$ mask. Based on these designs, each instance is represented by a total of 21 tokens (\ie 4 tokens for coordinates, 1 token for class, 16 tokens for mask), and we use \texttt{[INS]} as the task token of instance segmentation, as shown in Figure~\ref{fig:insseg_seq}. We append the meaningless zero mask for the noise box and do not add loss on those mask tokens during training.

\subsection{Soft Token}
In a typical auto-regressive prediction procedure, the token with the maximal predicted probability is selected as the output, and used its embedding as the input to the decoder for the next prediction step. This approach is called \emph{hard-inference}. 
However, since the tokens learned by VQ-VAE are not completely independent of each other, the correlation between the tokens may affect the token prediction accuracy, making the hard inference probably not optimal. To leverage the correlation, a \emph{soft token} technique is presented in the inference: instead of directly using the embedding of a single token, the soft token is the weighted averaged embedding of different tokens by their prediction probability. In addition to being applied in task-solver to predict the next token more accurately, the same idea can also be used in detokenizor to get better reconstruction results. 

Furthermore, the soft token results in the embedding space being spanned to an interpolable continuous space. Therefore, we can introduce an auxiliary loss that learns the task-specific output targets in an end-to-end manner by backing from the detokenzior output to the task-solver input. 

We examined the soft token in both instance segmentation and depth estimation. It can consistently improve the performance without any additional computational cost, which is a \emph{free-lunch} technique in inference.

\subsection{Mask Augmentation in Depth Estimation}
\begin{figure}
    \centering
    \includegraphics[width=1\linewidth]{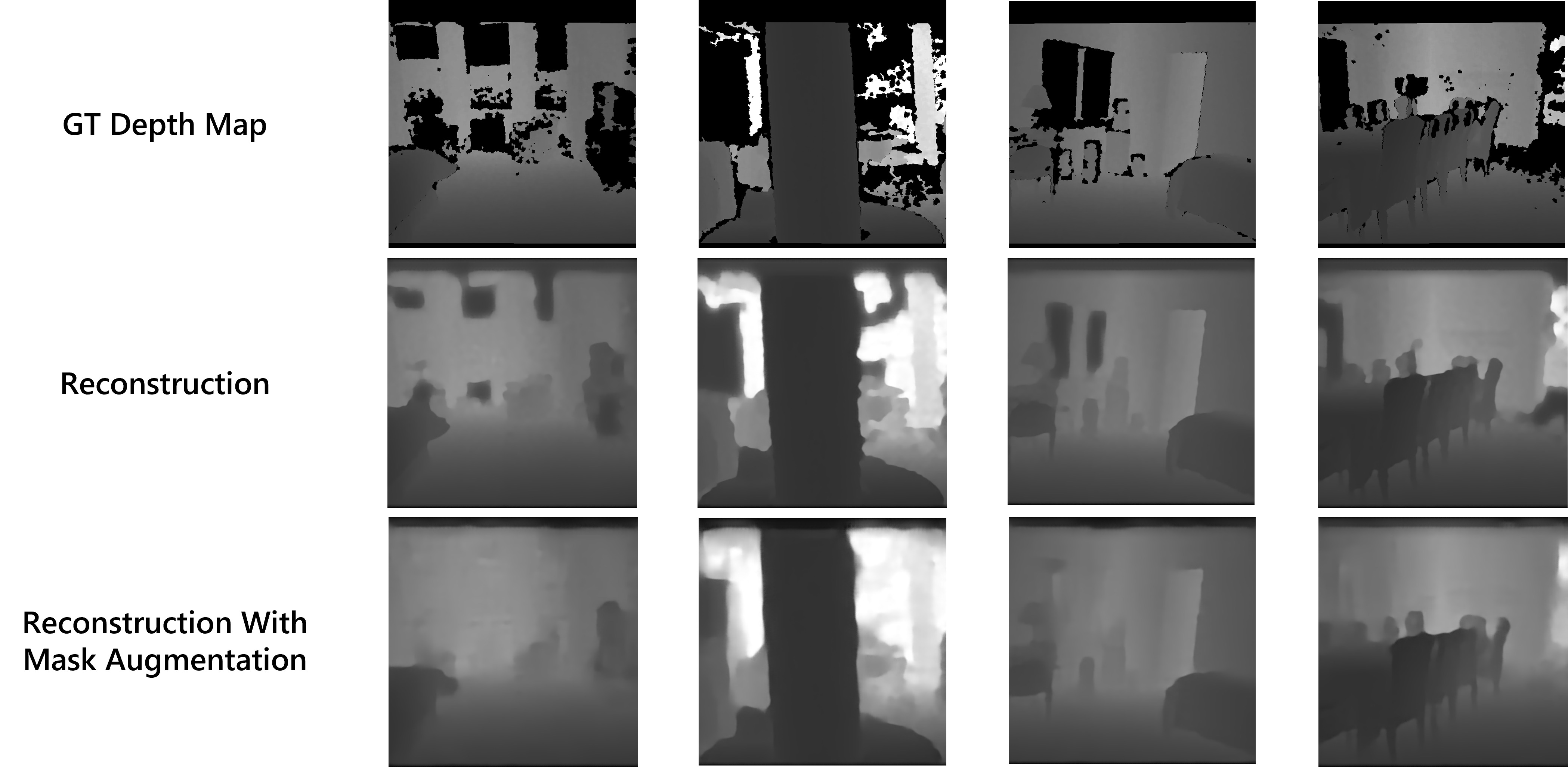}
    \caption{There are some corrupted regions (black regions/pixels) in the GT depth map. While we have ignored these regions in training VQ-VAE as well, the reconstructed regions are still abnormal, which is reflected in the shadows in reconstruction results. This phenomenon can be alleviated by adding masked augmentation.}
    \label{fig:aug_recons}
\end{figure}

\begin{figure}
    \centering
    \includegraphics[width=1\linewidth]{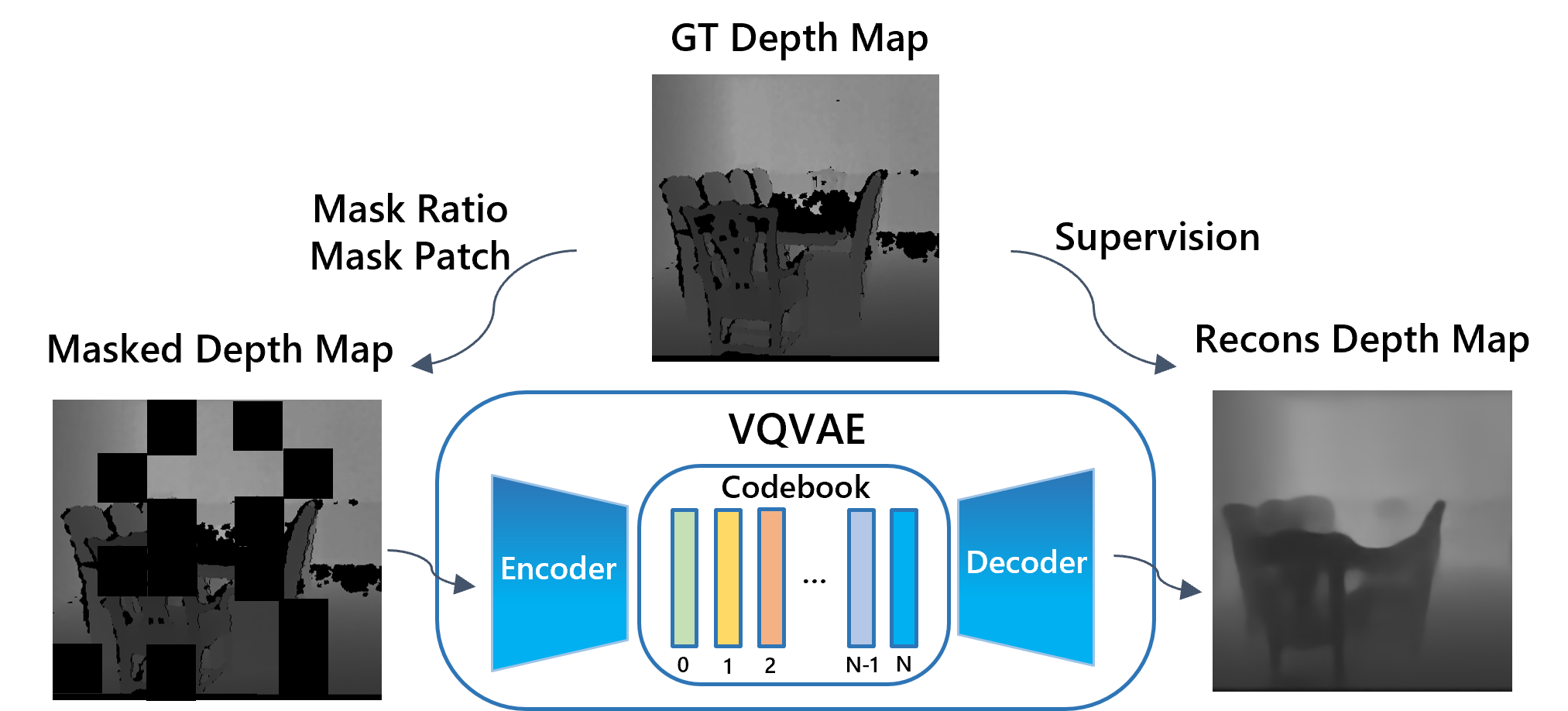}
    \caption{Illustration of our masked augmentation. In order to make the tokenizer have a stronger complement capability, we add random mask noise under a certain mask ratio and patch size to the original GT depth maps as inputs. But we still use the GT depth maps to apply reconstruction loss.}
    \label{fig:mask_aug_method}
\end{figure}

The ground-truth depth maps in the depth estimation dataset often have some corrupted regions that are not annotated with depth information. In the conventional depth estimation frameworks, these regions are ignored during training. However, the same solution cannot be applied to our framework. There are two challenges: First, while we have ignored these regions in training VQ-VAE as well, the reconstructed regions are still abnormal (see Figure \ref{fig:aug_recons}) and further affect the training of the task-solver and make the final result also have many artifacts; Second, a token predicted by VQ-VAE corresponds to a $32^2$ patch, which may contain both normal and corrupted pixels. Therefore, it is hard to deal with this issue by ignoring the tokens.

As shown in Figure \ref{fig:mask_aug_method}, we present to introduce mask augmentation in the training of VQ-VAE to alleviate this challenge. Specifically, we randomly mask some regions in the input depth images and then use their original depth information as supervision. In this way, the VQ-VAE can complete/recover some corrupted regions with reasonable results. Figure \ref{fig:aug_recons} shows the visualization.  In Table \ref{tab:depth_mask}, we notice that applying mask augmentation can improve the performance of depth estimation.

\section{Experiments}
\subsection{Tasks and Datasets}
To examine the generalizability of our framework, we choose depth estimation and instance segmentation, which are two tasks with very different output spaces.

\paragraph{Instance Segmentation} The instance segmentation requires predicting the location, the class, and the mask of each instance. The COCO2017 benchmark is one of the most challenging datasets for this task. It consists of 117K training images, 5K validation images, and 41K test images. A total of 80 classes annotation are provided. In our experiments, we follow the common setting of previous works~\cite{he2017mask,yang2020dense,xie2020polarmask} that report the performance on the validation set for comparison.

\paragraph{Depth Estimation} 
Depth estimation is a fundamental problem in computer vision, which requires estimating the depth for each pixel. Unlike segmentation whose output is a binary mask, the depth map is a floating point image. In this work, we use the NYUv2 Depth dataset, which consists of 24K training images and 654 validation images, and the RMSE is used as the major metric.

\subsection{Implementation Details}
We train two separate VQ-VAE for depth estimation and instance segmentation. For depth estimation, the input image size used in depth is $480^2$, with a batch size of 8. The Adam optimizer is used with the base learning rate of 3e-4, $\beta_1=0.9$, $\beta_2=0.999$. An exponential learning rate schedule is applied with the learning rate decay of 0.98 and a total of 20 training epochs. For instance segmentation. The input image size is $64^2$ with a batch size of 512. The Adam optimizer is used with the base learning rate of 3e-4, $\beta_1=0.9$, $\beta_2=0.999$. A cosine learning rate schedule is applied with a total of 20 training epochs. By default, the EMA model update technique is used for all VQ-VAE models.

For the task-solver, we adopt the auto-regressive encoder-decoder architecture, which is similar to Pix2Seq\cite{chen2021pix2seq}. It consists of a backbone, 6 encoder layers, and 6 decoder layers. We use the SwinV2\cite{liu2022swin} as the backbone, which is pre-trained with SimMIM~\cite{xie2022simmim}.
Most experiments in the ablation study are separately trained on depth estimation and instance segmentation. 

In depth estimation, we use the AdamW optimizer with a base learning rate of 2e-4 and 1e-4, the weight decay of 0.05 and 0.075 for SwinV2-B and SwinV2-L, respectively. The $\beta_1$ and $\beta_2$ are set to 0.9 and 0.999, and drop path rate is set to 0.1. The total training length is 25 epochs with the batch size of 24. The step learning rate schedule is used and the learning rate dropped to 2e-5 at the 18th epoch. For data augmentation, the random cropping of $480^2$ and horizontal flip with probability 0.5 are employed. We also append random brightness contrast, random gamma, and hue saturation value. 

Training of instance segmentation from scratch is expensive because of the long sequence length. To reduce the cost, we first train an object detection model and then fine-tuning on instance segmentation. For object detection pretraining, the AdamW optimizer with a base learning rate of 1e-3, a weight decay of 0.05, a drop path rate of 0.3, and a layer decay of 0.85, linear decay learning rate scheduler are applied, and a total of 100 training epochs with a batch size of 128 are performed. For instance segmentation fine-tuning, we only initialized the backbone and encoder with detection pre-trained model, and randomly initialized the decoder. In addition, AdamW optimizer with a base learning rate of 1e-4, a weight decay of 0.05, and a layer decay of 0.85, linear decay learning rate scheduler are applied, and the total training length is 50 epochs with a batch size of 16. Large-scale jittering with the range of $0.1$ to $3.0$ and crop size of $640^2$ are used for both object detection and instance segmentation. $\beta_1=0.9$, $\beta_2=0.999$ are used for AdamW in all experiments.

\subsection{Ablation Study}
We ablate the key design choices and techniques in this section. By default, we train the model separately for each task in the ablation study for better illustration, and SwinV2-B is used as the default backbone. For depth estimation experiments, a VQ-VAE with codebook size of 128, downsample rate of 32 and mask ratio of 0.5 is used by default. For instance segmentation, the codebook size is 128 and the downsample rate is 16. If not specified, we use the soft token but do not apply auxiliary loss for all ablation experiments.

\begin{table}[h]
\small
\caption{Ablation study on codebook size of VQ-VAE on depth and instance segmentation in reconstruction.}
\centering
\addtolength{\tabcolsep}{-4.9pt}
  \begin{tabular}{c|c|c|c}
    \hline

   \multirow{2}{*}{Width} & \multirow{2}{*}{\#tokens} & Depth & Instance Seg.  \\
  \cline{3-4}
  & & RMSE & Mask mAP \\
  \hline
  1.0$\times$ & 64 & 0.1025 & 88.94 \\
  1.0$\times$ & 128 & 0.0966 & 89.34 \\
  1.0$\times$ & 256 & \textbf{0.0902} & \textbf{90.22} \\
  \hline
  \end{tabular}
  \label{tab:codebook_VQ-VAE}
\vspace{-5pt}
\end{table}

\begin{table}[h]
\small
\caption{Ablation study on codebook size of VQ-VAE on depth and instance segmentation in task-solver. }
\centering
\addtolength{\tabcolsep}{-4.9pt}
  \begin{tabular}{c|c|c|c|c}
  \hline
   \multirow{2}{*}{Width} & \multirow{2}{*}{\#tokens} & Depth & \multicolumn{2}{c}{Instance Seg.}  \\
  \cline{3-5}
  & & RMSE & Box mAP & Mask mAP \\
  \hline
  1.0$\times$ & 64 & 0.3090 & 43.5 & 33.0 \\
  1.0$\times$ & 128 & \textbf{0.3080} & \textbf{43.6} & 33.2 \\
  1.0$\times$ & 256 & 0.3119 & 43.4 & \textbf{33.4} \\
  \hline
  \end{tabular}
  \label{tab:codebook_solver}
\vspace{-5pt}
\end{table}

\begin{table}[h]
\small
\caption{Ablation study on the width of VQ-VAE on depth and instance segmentation in reconstruction.}
\centering
\addtolength{\tabcolsep}{-4.9pt}
  \begin{tabular}{c|c|c|c}
  \hline
   \multirow{2}{*}{Width} & \multirow{2}{*}{\#tokens} & Depth & Instance Seg.  \\
  \cline{3-4}
  & & RMSE & Mask mAP \\
  \hline
  0.5$\times$ & 128 & 0.1196 &  88.97\\
  1.0$\times$ & 128 & \textbf{0.0966} &  89.34\\
  2.0$\times$ & 128 & 0.1025 &  \textbf{90.92}\\
  \hline
  \end{tabular}
  \label{tab:dim_VQ-VAE}
\vspace{-5pt}
\end{table}

\begin{table}[h]
\small
\caption{Ablation study on the width of VQ-VAE on depth and instance segmentation in task-solver.}
\centering
\addtolength{\tabcolsep}{-4.9pt}
  \begin{tabular}{c|c|c|c|c}
    \hline
   \multirow{2}{*}{Width} & \multirow{2}{*}{\#tokens} & Depth & \multicolumn{2}{c}{Instance Seg.}  \\
  \cline{3-5}
  & & RMSE & Box mAP & Mask mAP \\
  \hline
  0.5$\times$ & 128 & 0.3127 & 43.4 & 33.1 \\
  1.0$\times$ & 128 & \textbf{0.3080} & \textbf{43.6} & \textbf{33.2} \\
  2.0$\times$ & 128 & 0.3124 & 43.2 & 33.1 \\
  \hline
  \end{tabular}
  \label{tab:dim_solver}
\vspace{-5pt}
\end{table}

\paragraph{Architecture of VQ-VAE} 

We study how different designs of VQ-VAE affect performance. We first evaluate the reconstruction performance of different codebook sizes. To more accurately and intuitively observe the reconstruction performance, our evaluation is performed on the validation set of different tasks and adopts mask mAP and RMSE as metrics. The results are shown in Table \ref{tab:codebook_VQ-VAE}. We find that although the large codebook size (\eg 256) benefits the reconstruction performance, the small codebook size (\eg 64) can also yield sufficiently good reconstruction performance. Further applying to the task-solver, we found different codebook size has little effect on final performance (see Table \ref{tab:codebook_solver}).

Using the same evaluation method, we study the effect of the width of VQ-VAE. Table \ref{tab:dim_VQ-VAE} shows the reconstruction performance and Table \ref{tab:dim_solver} shows the performance of applying to task-solver. Similar to the observation on codebook size, we find that the network width has little effect on the final performance.

The downsample ratio of VQ-VAE is another key that may affect network performance. We vary the downsample ratio in [8, 16, 32]. We note that the instance segmentation cannot support a downsample ratio of 8 because it results in too long sequences. Table \ref{tab:ds_VQ-VAE} shows the reconstruction performance, as the downsample ratio increases, the reconstruction performance gets worse, satisfying the intuition. However, we find that better reconstruction performance does not always lead to better performance when applying the VQ-VAE in task-solver. In Table \ref{tab:ds_solver}, the best performance is achieved at downsample ratio of 32. We explain this phenomenon is that a smaller downsample ratio facilitates reconstruction, but it also increases the length of the token sequence, which is detrimental to the task-solver.

\begin{table}[t]
\small
\caption{Ablation study on the downsample ratio of VQ-VAE on depth and instance segmentation in reconstruction.}
\centering
\addtolength{\tabcolsep}{-4.9pt}
  \begin{tabular}{c|c|c}
  \hline
   \multirow{2}{*}{Downsample Ratio} & Depth & Instance Seg. \\
   \cline{2-3}
   & RMSE & Mask mAP \\
  \hline
  32 & 0.0966 & 70.91 \\
  16 & 0.0696 & \textbf{89.34} \\
  8 & \textbf{0.0515} & - \\
  \hline
  \end{tabular}
  \label{tab:ds_VQ-VAE}
\vspace{-5pt}
\end{table}

\begin{table}[t]
\small
\caption{Ablation study on the downsample ratio of VQ-VAE on depth and instance segmentation in task-solver.}
\centering
\addtolength{\tabcolsep}{-4.9pt}
  \begin{tabular}{c|c|c|c}
  \hline
   \multirow{2}{*}{Downsample Ratio} & Depth & \multicolumn{2}{c}{Instance Seg.} \\
   \cline{2-4}
   & RMSE & Box mAP & Mask mAP \\
  \hline
  32 & \textbf{0.3080} & 42.7 & 30.3 \\
  16 & 0.3304 & \textbf{43.6} & \textbf{33.2}  \\
  8 & 0.3514 & - & - \\
  \hline
  \end{tabular}
  \label{tab:ds_solver}
\vspace{-5pt}
\end{table}

\begin{table}[t]
\small
\centering
\caption{Ablation on the effectiveness of soft token.  }
\begin{tabular}{l|c|c|c}
\hline
\multirow{2}{*}{Description} & Depth & \multicolumn{2}{c}{Instance Seg.}\\
\cline{2-4}
& RMSE & box mAP & mask mAP\\
\hline
Baseline (hard-inf) & 0.3174 & 43.6 & 31.1 \\
\hline
On. task-solver & 0.3127 & 43.5 & 32.1\\
On. detokenizor & 0.3120 & 43.6 & 32.3 \\
On. both & 0.3080 & 43.6 & 33.2 \\
On. both + aux. loss & \textbf{0.3052} & \textbf{43.3} & \textbf{34.2} \\
\hline
\end{tabular}
\label{tab:soft_depth_ins}
\vspace{-5pt}
\end{table}

\paragraph{Soft Token}
To leverage the correlation between tokens, we introduce the soft token techniques, which can be used to improve the performance in the inference stage for free. We examined this technique in Table \ref{tab:soft_depth_ins}. Compared with the hard inference baseline, applying the soft token in task-solver and detokenzior alone can bring performance gains, and further performance improvement is achieved when used in two stages at the same time: the depth performance is improved by +0.009 RMSE and the instance segmentation is improved by +2.1 mAP. On top of it, adding the auxiliary loss on the output of detokenizor enlarges the gain to +0.012 RMSE and +3.1 mAP.

\begin{table}[h]
\small
\caption{Affects of mask augmentation in training VQ-VAE on depth. The patch size of all models is set to 16. }
\centering
\addtolength{\tabcolsep}{-4.9pt}
  \begin{tabular}{c|c|c}
  \hline
   Mask Ratio &  VQ-VAE & task-solver  \\
  \hline
  0.0 &  \textbf{0.0831} & {0.3105} \\
  0.3 &  0.0893 & {0.3093} \\
  0.5 & 0.0966 & \textbf{0.3080}\\
  0.7 &0.1196 & {0.3225}\\ 
  \hline
  \end{tabular}
  \label{tab:depth_mask}
\vspace{-5pt}
\end{table}

\paragraph{Mask Augmentation}
We evaluate the effectiveness of mask augmentation in depth estimation. The different mask ratios vary from 0.3 to 0.7 are used, and Table \ref{tab:depth_mask} shows the results. The best performance is achieved by using the mask ratio of 0.5, which is +0.003 better than the baseline.

\begin{figure*}
    \centering
    \includegraphics[width=0.96\linewidth]{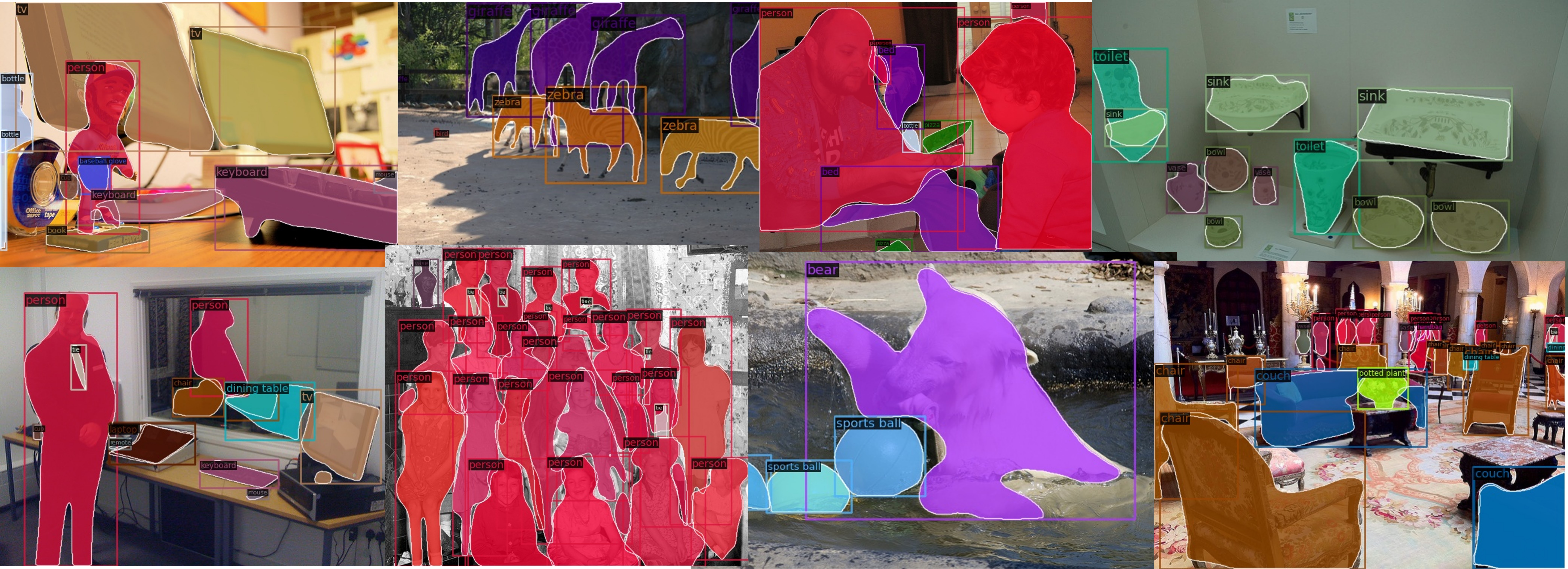}
    \caption{Visualization on instance segmentation task of our method. }
    \label{fig:insseg_vis}
\end{figure*}

\begin{figure*}
    \centering
    \includegraphics[width=0.96\linewidth]{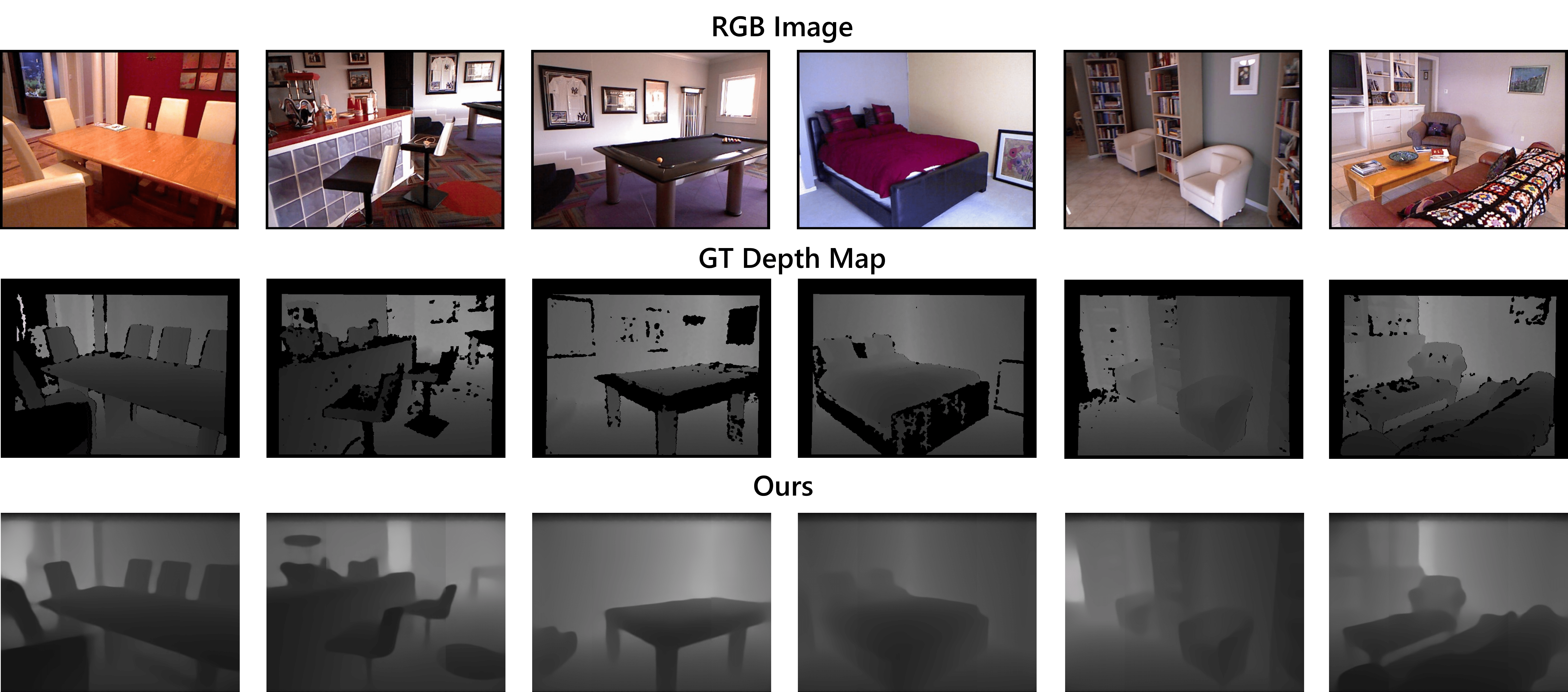}
    \caption{Visualization of depth estimation task. With the well-designed structure and techniques, our model is capable for this task.}
    \label{fig:depth_vis}
\end{figure*}

\renewcommand{\arraystretch}{1.3}
\begin{table}[t]
		\caption{Results of monocular depth estimation task on NYUv2~\cite{nathan2012nyuv2}. Both AiT and AiT-P are our methods, where AiT indicates auto-regressive prediction, and AiT-P indicates parallel prediction.}
		\vspace{-0.3cm}
		\centering\setlength{\tabcolsep}{3.35pt}
		\label{table:nyuv2}
		\footnotesize
		\begin{tabular}{l|cccccc}
			\hline
			Method & $\operatorname{RMSE}\downarrow$ & $\delta_1\uparrow$ & $\delta_2\uparrow$ & $\delta_3\uparrow$ & $\operatorname{REL}\downarrow$ &  $\operatorname{log10}\downarrow$ \\
			\hline
			DORN~\cite{Fu18dorn} & 0.509 & 0.828 & 0.965 & 0.992 & 0.115  & 0.051  \\
            BTS~\cite{Lee19bts} & 0.392 & 0.885 & 0.978 & 0.995 & 0.110 & 0.047\\
            AdaBins~\cite{Bhat21adabins} & 0.364 & 0.903 & 0.984 & 0.997 & 0.103 & 0.044\\
			DPT~\cite{Ranftl21DPT} & 0.357 & 0.904 & 0.988 & 0.998 & 0.110 & 0.045\\
			LocalBins~\cite{Bhat22localbins} & 0.357 & 0.907 & 0.987 & 0.998 & 0.099 & 0.042\\
			P3Depth~\cite{patil22p3depth} & 0.356 & 0.898 & 0.981 & 0.996 & 0.104  & 0.043\\
			BinsFormer~\cite{li22binsformer} & 0.339 & 0.921 & 0.989 & 0.998 & 0.096 & 0.041\\
			NeWCRFs~\cite{yuan22newcrf} & 0.334 & 0.922 & 0.992 & 0.998 & 0.095 & 0.041\\
			BinsFormer~\cite{li22binsformer} & 0.330 & 0.925 & 0.989 & 0.997 & 0.094 & 0.040\\
            SwinV2-B~\cite{xie2022revealing} & 0.303 & 0.938 & 0.992 & 0.998 & 0.086 & 0.037 \\
            SwinV2-L~\cite{xie2022revealing} & 0.287 & 0.949 & 0.994 & 0.999 & 0.083 & 0.035 \\
            \hline
            UViM\cite{kolesnikov2022uvim} & 0.467 & - & - & - & - & - \\
             Unified-IO$_{XL}$\cite{lu2022unified} & 0.385 & - & - & - & - & -  \\
			\hline
            AiT (SwinV2-B) & 0.305 & 0.934 & 0.991 & 0.998 & 0.087 & 0.037 \\
            AiT (SwinV2-L) & 0.284 & 0.949 & 0.993 & 0.999 & 0.079 & 0.034 \\
            \hline
            AiT-P (SwinV2-B) & 0.301 & 0.940 & 0.992 & 0.998 & 0.085 & 0.036 \\
            AiT-P (SwinV2-L) & \textbf{0.275} & \textbf{0.954} & \textbf{0.994} & \textbf{0.999} & \textbf{0.076} & \textbf{0.033} \\ 
            \makecell{AiT-P (SwinV2-L) \\ w/o soft token}  & 0.282 & 0.951 &
            0.994 & 0.999 & 0.080 & 0.034 \\ 
			\hline
		\end{tabular}
		\vspace{-.3cm}
	\end{table}

\subsection{Preliminary Study on Parallel Prediction}

In previous sections, we mainly demonstrate the use of auto-regressive models to unify various visual tasks. Note the bi-directional or even qua-directional natural may encourage parallel decoding approaches. Therefore, we make some preliminary studies on the parallel decoder instead of the auto-regressive decoder. 

As parallel decoder has been prevalent in object detection and instance segmentation, \ie DETR~\cite{carion2020end} and Mask2Former\cite{cheng2022masked}, we mainly studied the parallel decoder for the depth estimation problem. As shown in Table~\ref{table:nyuv2}, the parallel decoder performs better than the auto-regressive counterpart by around 0.01 RMSE (0.275 \vs 0.284). This indicates a strong potential than the auto-regressive models.

Our techniques such as soft token also benefit the parallel decoder, as shown in Table~\ref{table:nyuv2}. This implies the generality of the proposed techniques.

\subsection{Comparison with Other Unified Frameworks and State-of-the-arts in Depth Estimation}

UViM and Unified-IO are the most relevant works to ours. We compare the performance with these methods on the overlap task, \ie NYUv2 depth estimation. The results are shown in Table.~\ref{table:nyuv2}. 
Our auto-regressive approach achieves 0.284 RMSE, which is 0.183 and 0.101 better than UViM and Unified-IO. Moreover, our parallel approach further improves the performance to 0.275 RMSE. This result surpasses previous state-of-the-arts by 0.012 RMSE. While UViM and Unified-IO mainly conceptually propose unified frameworks for various visual tasks, we push more solid steps through in-depth study of the general visual task-solver.

 \subsection{One Model for Multiple Tasks}
We train the instance segmentation and depth jointly using a shared task-solver. Table~\ref{tab:joint} shows that the joint training with shared model weights has marginal performance gradation compared to using separate task solvers.

\begin{table}[t]
\small
\caption{Joint training of depth estimation and instance segmentation using a single task-solver. The performance of joint training is slightly worse compared to using separate task-solvers for each task.}
\centering
\addtolength{\tabcolsep}{-4.9pt}
  \begin{tabular}{c|c|c|c}
  \hline
   \multirow{2}{*}{Description.} & Depth & \multicolumn{2}{c}{Instance Seg.} \\
   \cline{2-4}
    & RMSE & Box mAP & Mask mAP \\
  \hline
  separate training & \textbf{0.3052} & \textbf{43.3} & \textbf{34.2} \\
  joint training & 0.3103 & 42.2 & 34.1 \\
  \hline
  \end{tabular}
  \label{tab:joint}
\vspace{-5pt}
\end{table}

\section{Conclusion}
In this work, we investigate the unification of output spaces for various vision tasks by a set of visual tokens, and further develop a unified auto-regressive encoder-decoder model. Two new techniques are proposed which take the particularity of visual tasks into account to improve the system: 1) Soft token can leverage the correlation between tokens to improve performance in the inference stage and enables end-to-end learning for the final visual targets; 2) Mask augmentation is used to alleviate the issue of corrupted/undefined areas of visual tasks, \ie depth estimation. With these two techniques, our general method set a new state-of-the-art on the NYUv2 depth dataset, as well as competitive accuracy on the COCO 2017 object detection and instance segmentation benchmark. We also conducted preliminary studies on the parallel decoder which is show promising results on the depth estimation problem.

\appendix

\section{Ablation on the Depths of VQ-VAE}
In Table~\ref{tab:dim_VQ-VAE} and Table~\ref{tab:dim_solver}, we have studied how different widths of VQ-VAE affect the performance. In Table~\ref{tab:nresblocks}, we further examine the effect of depth, \ie different number of residual blocks in VQ-VAE. The results show that increasing the number of residual blocks cannot bring more improvements and we find greater results variance when training tokenizer with deeper networks.

\begin{table}[t]
\small
\caption{Ablation study on increasing the number of residual blocks on the 32$\times$ downsampling setting of depth estimation.}
\centering
\addtolength{\tabcolsep}{-4.9pt}
  \begin{tabular}{c|c|c}
  \hline
   \multirow{2}{*}{\#Resblock} & Tokenizor & Task-solver \\
   \cline{2-3}
   & RMSE & RMSE \\
  \hline
  2 & \textbf{0.0966} & \textbf{0.3080} \\
  3 & 0.1055 & 0.3123 \\
  4 & 0.1082 & 0.3136 \\
  5 & 0.1033 & 0.3118 \\
  \hline
  \end{tabular}
  \label{tab:nresblocks}
\vspace{-5pt}
\end{table}

\section{VQ-VAE \vs Interpolation Tokenizer}
We use a lightweight VQ-VAE as the tokenizer to reduce the resolution and computation for the task-solver. A intuitive baseline to perform down-sampling is the interpolation method. For example, in instance segmentation, we can use a nearest-neighbor interpolation method to up-sample the binary masks. For depth estimation, we use a bilinear interpolation method and discretize the floating numbers between $0$ to $10$ by 128 bins. We select the same input and target resolution as our VQ-VAE tokenizer for a fair comparison. As shown in Table~\ref{tab:interpolate}, the tokenizer using nearest-neighbor interpolation is much worse than our VQ-VAE method, showing the effectiveness of using VQ-VAE to represent the task output space.

\begin{table}[t]
\small
\caption{Comparison of VQ-VAE and interpolation as the tokenizer.}
\centering
\addtolength{\tabcolsep}{-4.9pt}
  \begin{tabular}{c|c|c|c}
  \hline
   \multirow{2}{*}{Description.} & Depth & \multicolumn{2}{c}{Instance Seg.} \\
   \cline{2-4}
   & RMSE & Box mAP & Mask mAP \\
  \hline
  Ours & \textbf{0.3080} & 43.6 & \textbf{33.2} \\
  interpolation & 0.7544 & \textbf{43.8} &  4.2 \\
  \hline
  \end{tabular}
  \label{tab:interpolate}
\vspace{-5pt}
\end{table}

\section{Details for Joint Training}

The AdamW optimizer with a base learning rate of 8e-4, $\alpha_1=0.9$, $\alpha_2=0.999$, a weight decay of 0.05, and a layer decay of 0.85, a drop path rate of 0.3, and the linear decay learning rate scheduler are applied. We train the model with 300k iterations and batch size 64. The decoder loss weights of 5.0 and 1.0 are used for instance segmentation and depth estimation respectively. A depth loss weight of 0.2 is used. 
We process these batch data individually \ie different tasks do task-specific augmentation and generate their discrete training token sequence. 

{\small
\bibliographystyle{ieee_fullname}
\bibliography{egbib, depth}
}

\end{document}